\documentclass[letterpaper, 10 pt, conference]{ieeeconf}  

\IEEEoverridecommandlockouts                              

\usepackage[noadjust]{cite}
\usepackage{booktabs}
\usepackage{float}
\usepackage{amsmath}
\usepackage{amsfonts}
\usepackage{graphicx}
\usepackage{subcaption}
\let\labelindent\relax
\usepackage{enumitem}
\usepackage[dvipsnames,table]{xcolor}
\usepackage{textcomp, gensymb}
\usepackage{hyperref}
\usepackage{lipsum}
\usepackage{balance}
\usepackage[switch]{lineno}
\usepackage{mathtools}
\usepackage[linesnumbered,ruled,vlined,english,onelanguage]{algorithm2e}
\usepackage{lipsum}
\usepackage{multirow}
\usepackage{multicol}
\usepackage{amssymb} 
\usepackage{makecell}
\usepackage{titlesec}

\title{\LARGE \bf
SENT-Map: Semantically Enhanced Topological Maps\\with Foundation Models
}

\author{Raj Surya Rajendran Kathirvel$^{1}$, Zach Chavis$^{2}$, Stephen J. Guy$^{2}$, and Karthik Desingh$^{1, 2}$\\
\thanks{$^{1}$Minnesota Robotics Institute (MnRI), and $^{2}$Department of Computer Science and Engineering (CS\&E), University of Minnesota, Minneapolis, MN 55414 US.
{\tt\small(rajen064|chavi014|sjguy|kdesingh)@umn.edu}}%
}

\let\oldtwocolumn\twocolumn
\renewcommand\twocolumn[1][]{%
    \oldtwocolumn[{#1}{
           \centering
           \includegraphics[width=1.0\textwidth]{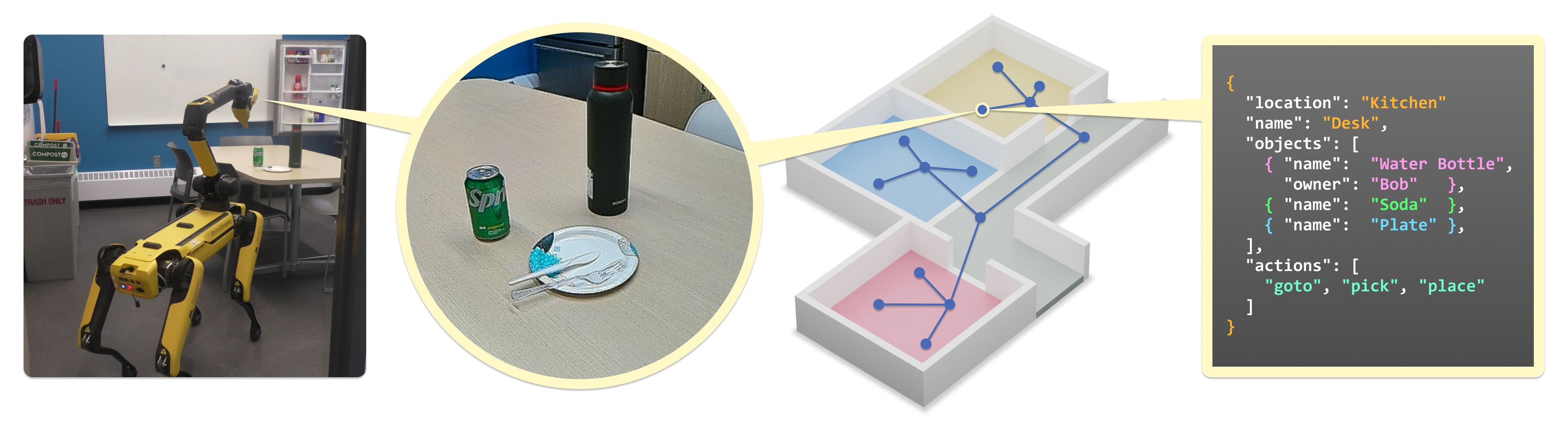}

        \begin{minipage}{0.97\textwidth}
        \begin{minipage}{0.225\textwidth}
            \raggedright
            \textbf{(a)} Operator's perspective
        \end{minipage}\hfill
        \begin{minipage}{0.225\textwidth}
            \raggedright
            \textbf{(b)} Robot's perspective
        \end{minipage}\hfill
        \begin{minipage}{0.225\textwidth}
            \raggedright
            \textbf{(c)} Topological map
        \end{minipage}\hfill
        \begin{minipage}{0.225\textwidth}
            \raggedright
            \textbf{(d)} Semantic node JSON
        \end{minipage}
        \end{minipage}
        
           \captionof{figure}
           {\small{\textbf{SENT-Map overview.} (\textbf{a}) An operator guides a robot throughout an environment, (\textbf{b}) stopping at task-relevant locations to take a snapshot, (\textbf{c}) which is tagged with a location and appended to a graph, and (\textbf{d}) fed to a foundation model alongside human annotation to generate a JSON description of the semantic context.}}
           \label{fig:teaser}
    }]
}

\begin{document}

\maketitle
    
\section{Introduction}
We introduce  SENT-Map, a semantically enhanced topological map for representing indoor environments, designed to support autonomous robot navigation and manipulation by leveraging advancements in foundational models  (FMs).  The general semantic information and planning capabilities found in modern FMs provide exciting potential for robots operating in complex human environments to have powerful capabilities in unconstrained, open-world environments. However, these FMs also come with notable risks related to hallucination, false confidence, and other assurance-related issues. We propose to address these issues by grounding plans from FMs in the real-world locations as represented through a topological map~\cite{KUIPERS199147}. Additionally, we rely on both vision-language models (VLMs) and human operators to further enhance these topological maps with semantic information, allowing a robot to build robust plans to account for its navigation capabilities, manipulation affordances, and other semantic information about objects, rooms, and people in the robot's environment. By incorporating these resulting SENT-Maps with a Large-Language Model (LLM) for planning, the resulting system is able to execute a variety of navigation-manipulation tasks in complex environments specified through natural language commands.

At a high level, the proposed framework operates in two stages: a mapping stage and a planning/execution stage. The mapping stage begins with a navigational map build through an environmental walkthrough followed by the use of an FM to construct the SENT-Map. The SENT-Map construction process is human-guided in two ways: first, a human operator is responsible for walking the robot through the environment, and second, the operator highlights various semantically interesting locations and objects.  
The result of the mapping stage is a structured SENT-Map in JSON format, representing the environment as a topological graph with nodes containing robot affordances, navigability links, and any additional environmental information useful for mobile manipulation tasks such as people's locations or objects' ownership.  Critically, these node-based maps can be easily visualized, checked, and edited by a human for not only correctness, but also supplementary semantic information not privy to FMs. The planning and execution stage is executed by a Planning FM, in this case an LLM, which takes the SENT-Map in JSON format, a description of the robot's known skills, and a natural language command as input to generate and execute task plans. These actions generated by the FM can be limited at plan creation time to those allowed by the SENT-Map, eliminating the possibility of hallucinations at plannign time related to affordances or capabilities.

In summary, our contributions are as follows:
\begin{itemize}
\item SENT-Map, a Semantically-Enhanced Topological Map in human-interpretable JSON for autonomous robot navigation and manipulation
\item A framework for constructing and planning over SENT-Maps using existing foundation models.
\item Experimental results showing SENT-Maps improve FM planning success even on locally-deployable FMs.
\end{itemize} 

\section{Related Works}

Liu et al. proposed FM-fusion~\cite{FM-Fusion}, an instance-aware semantic mapping framework combining vision-language models with SLAM for camera pose estimation. While it enables open-set labeling and dense segmentation, it suffers from high computational demands, lacks embodiment-specific affordance reasoning, and provides no assurances for task execution. For building SENT-Maps, we utilize FMs to generate a map in JSON, which is human-editable to further enhance or correct the map to allow for assured task execution. 


\par Object-centric mapping approaches~\cite{Slam++, SemanticMapping} integrate rich semantic information about objects into maps, making them highly effective for manipulation tasks. While these methods excel in supporting intricate manipulation scenarios, current foundational models are not yet sufficiently advanced for reliable open-set 3D reconstruction or pose estimation, though promising progress has been made~\cite{hughes2022hydra}. 
Our proposed framework seeks to address this limitation by leveraging FM models for 2D visual semantic understanding combined with natural language utterances to construct sparse representations, as seen in SENT-Maps. 

\par Several works leverage FMs to semantically represent environments for visual navigation. CLIP-Fields~\cite{ClipField} maps 3D spatial locations to high-dimensional feature vectors embedding CLIP-based language and visual features, while NLMap-SayCan~\cite{OpenVocab} uses a 2D grid-based map with discrete object representations derived from a region proposal network and VLM features. VLMaps~\cite{vlmaps} represents fixed object sets in a 2D grid as top-down projections, enabling spatial goal navigation using language commands. Techniques like 3D-LLMs~\cite{3dllm} and the real-time OpenFusion~\cite{openfusion} offer open-vocabulary 3D mapping and queryable scene representations using RGB-D data. While these methods are effective, they represent semantic information in feature space, making their maps non-verifiable by humans and limiting their ability to incorporate affordances. In contrast, our proposed framework uses FM during mapping phase to achieve open-set semantic enhancement, enabling the creation of human-verifiable editable maps.


\par Graph-based methods represent scenes as 3D graphs, embedding geometric and semantic information into nodes for high-level task planning. Concept Graphs~\cite{conceptgraph} constructs 3D graphs by leveraging 2D foundation models and fusing their outputs into 3D through multi-view association, enabling open-vocabulary representations without extensive 3D datasets. CLIO~\cite{clio} builds task-driven 3D object-centric maps, clustering object primitives into semantic regions based on task specifications. While these methods create object-centric topological maps, they require task specifications during map creation, and are not designed for editability, limiting their flexibility and usability. by enabling open-set semantic enhancement during the mapping phase and task specification through natural language interaction in the execution phase, thereby improving adaptability and task assurance.

\section{Semantically Enhanced Topological Map}

\begin{figure*}[th!]
     \centering
    \vspace*{0.5em}
    \includegraphics[width=1.0\textwidth]{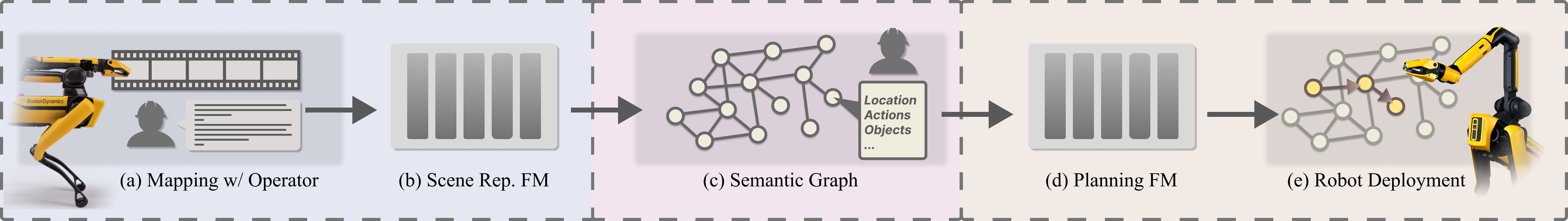}
     \caption{\small{\textbf{SENT-Map Framework.} (\textbf{a}) An operator defines a map alongside a robot. (\textbf{b}) Images and operator prompt are given to a Scene Representation FM, which outputs a node in JSON. (\textbf{c}) A collection of nodes defines our semantic graph. Due to the interpretability of JSON, the operator is free to make additions or corrections within the JSON. (\textbf{d}) The full JSON graph is fed to a planning FM alongside a query, and a skill sequence is output. (\textbf{e}) The robot then executes the skill sequence within the environment.}}
    \label{fig:framework}
\end{figure*}

\subsection{Problem Definition}
Consider a topological map $\mathcal{M}$ represented as a graph $\mathcal{M} = G(V, E)$, where $V$ denotes the nodes the robot can navigate to and $E$ denotes the navigation actions it can perform to move between the vertices. The SENT-Map embeds additional information into a subset of the vertices $V_{SE} \subseteq V$, as illustrated in Fig.~\ref{fig:teaser}. Our framework has two phases and hence two problems: one, constructing the map $\mathcal{M}$, where a human walks the robot in the indoor environment, facilitating the capture of RGB observations $I_v$ of the semantic node locations $v$, which are then passed through an FM (Scene Representation FM) $\mathcal{M} \leftarrow \{S_{FM}(I_v) \mid v \in V_{SE}\}$ to output the JSON-structured SENT-Map $\mathcal{M}$; and two, planning with the constructed $\mathcal{M}$, where given the SENT-Map and a natural language query, an FM (Planning FM) outputs the navigation and manipulation sequence to complete the task. Given a target node $v_{target}$, a current node $v_{current}$, and the map $\mathcal{M}$, we assume that there exists a path planner that gives the shortest path the robot can successfully execute to reach $v_{target}$.


\subsection{Scene Representation (SENT-Map)}
To enable reasoning by language models over physical environments, the SENT-Map represents the environment as a JSON-structured topological graph. It encodes spatial and semantic information in a hierarchical format, grounded in the robot’s navigation and interaction capabilities. At the top level, the SENT-Map consists of navigation nodes, which represent traversable waypoints in free space. Each navigation node specifies its connectivity to other nodes via directed edges, forming the graph structure $\mathcal{M} = G(V, E)$ as shown in Fig~\ref{fig:framework} (c). Each navigational node has the potential to be a semantic node, representing nearby stationary entities such as fridges, drawers, cabinets, tables, or desks that serve as potential locations for robot interaction. As entities represented by semantic nodes may require additional context (e.g., a fridge must be opened to retrieve its contents), the semantic node can indicate how the entity's state can be changed by the robot through manipulation. These semantic nodes may also contain objects, which are movable, graspable entities such as mugs, tissue boxes, bottles, or cans; these objects are targets for high-level tasks such as ``get," ``move," or ``clean." Each semantic node may further contain additional metadata relevant for downstream tasks, such as tagging individual objects with ownership, e.g. ``Bob's mug." This structured and interpretable representation enables foundational models to reason over physical spaces using natural language, while supporting efficient navigation and task planning.

\subsection{Foundation Models Maps with Human Grounding}
To construct a SENT-Map, an operator guides a robot around an environment as shown in Fig~\ref{fig:teaser}~(a). As the robot moves, navigational nodes are created, memorizing the free-space of the environment. During mapping, the operator stops at desired points of interest, which become semantic nodes. To construct the semantic node, the robot first takes an RGB snapshot using onboard cameras as shown in Fig~\ref{fig:teaser}~(b). The image is then passed to a Vision-FM along with a prompt defining the JSON template, i.e. the structure of a semantic node, as shown in Fig~\ref{fig:teaser}~(d). After mapping, the collection of semantic nodes is represented as a Scene JSON, a compressed-textual representation of the map.



As the JSON format is human-interpretable, during this mapping phase the user can also modify/rectify the JSON or add other additional information such as ownership tags (``this is person A's mug"), associated with the semantic node or any of the objects it contains to enrich the SENT-Map. Any hallucination or incorrect inference made by the mapping FM can be corrected by the operator before the planning stage. Hence, the JSON used during the planning stage will be representative of the ground truth. An example segment from a SENT-Map JSON file is shown in Fig~\ref{fig:framework} (c).


\subsection{Planning with Foundation Models}
For planning, we employ a text-only foundation model as a Planning FM, which is tasked with determining the series of steps that satisfy a task given the Scene JSON and user-defined query as shown in Fig~\ref{fig:framework}~(d). Following~\cite{bumble}, we ground the output actions to a skill API, which defines the actions the robot can take, and a description of the robot's physical constraints (i.e. single arm). The final prompt for the FM contains the scene, the skill API, the physical robot constraints, and the user query. Once the FM returns a plan, the robot executes the series of commands using the navigation nodes for global motion planning, and off-the-shelf methods for local collision avoidance and object manipulation.


\section{Experimental Results}

\begin{table*}[ht!]
    \centering
    \begin{tabular}{l|cccr|cccr}
    \toprule
     & \multicolumn{4}{c|}{Baseline} & \multicolumn{4}{c}{Semantic Enhancement} \\ \cline{2-9}
    \rule{0pt}{3ex}Model & Sponge & Coffee & Tissue & \textit{Average} & Sponge & Coffee & Tissue & \textit{Average} \\ \midrule

    \rowcolor{gray!10} 
    Gemma 3 27B & \checkmark & \texttimes & \checkmark &  66.7\% & \checkmark  &  \checkmark &  \checkmark &  100\% \\
    
    Gemini 2.0 Flash & $\varnothing$ & $\varnothing$ & $\varnothing$ & 0.0\% & \checkmark & \checkmark & \checkmark & 100\%\\
    
    \rowcolor{gray!10} 

    Llama 3.1 8B & \texttimes & \texttimes & \checkmark & 33.3\% & \checkmark & \checkmark & \checkmark & 100\%\\
    
    Llama 3.1 405B & \texttimes & \texttimes & \checkmark & 33.3\% & \checkmark & \checkmark & \checkmark & 100\%\\

    \rowcolor{gray!10} 
    GPT 4o mini & \texttimes & \texttimes & \checkmark & 33.3\% & \checkmark & \checkmark & \checkmark & 100\%\\
    
    GPT o3 mini & \checkmark & \checkmark & \texttimes & 66.7\% & \checkmark & \checkmark & \checkmark & 100\%\\
   
    \midrule
    \textit{Average} &  \multicolumn{4}{c|}{38.9\%} &  \multicolumn{4}{c}{100\%}\\
    \bottomrule
    \end{tabular}
    \caption{\small{Task success across several LLMs.} A ``\checkmark'' denotes task success, an ``\texttimes'' denotes task failure, and a ``$\varnothing$'' denotes the model's refusal to output a solution due to requiring additional context. }
    \label{tab:task_success}
\end{table*}

\begin{table}[ht!]
    \centering
    \begin{tabular}{l|cc|cc}
    \toprule
     & \multicolumn{2}{c|}{Baseline} & \multicolumn{2}{c}{Semantic Enhancement} \\ \cline{2-5}
    \rule{0pt}{3ex}Task& Direct & Indirect & Direct & Indirect\\ \midrule

    \rowcolor{gray!10} 
    Watch TV       & \texttimes & \texttimes &  \checkmark &  \checkmark \\
    
    Runny Nose     & \checkmark & \texttimes &  \checkmark &  \checkmark \\
    
    \rowcolor{gray!10}     
    Private listening     & \texttimes & \texttimes &  \checkmark &  \checkmark \\
    
    Sanitization     & \texttimes & \texttimes &  \checkmark &  \checkmark \\
    
    \rowcolor{gray!10}     
    Call a friend     & \texttimes & \texttimes &  \checkmark &  \checkmark \\
    
    Flavor Coffee       & \checkmark & \texttimes &  \checkmark &  
    \checkmark \\
    \midrule
    & \multicolumn{2}{c|}{SE} & \multicolumn{2}{c}{SE + Ownership} \\ \midrule
    \rowcolor{gray!10}  
    Store Bob's leftovers & \checkmark & \checkmark &  \checkmark &  \checkmark\\
    Get Bob his drink & \checkmark & \texttimes & \checkmark & \checkmark \\
    \rowcolor{gray!10}
    Bob's things to Alice & \checkmark & \texttimes & \checkmark &  \checkmark\\
    \bottomrule
    \end{tabular}
    \caption{\small{Direct-query and indirect-query task success for small foundation model.} Gemma 3 27B was prompted with two types of queries, one directly asking for the objects, and one indirectly suggesting the object without naming it. Results indicate that semantic enhancement enables even a small FM to reason about complex tasks.}
    \label{tab:indirect}
\end{table}

To evaluate the impact of semantic enhancement for planning, we compare the performance on the SENT-Map environment represented in Fig~\ref{fig:qualitative}, consists of nine different semantic nodes defined over three major zones (office, lounge and kitchen). Each room has various items for a total of 23 objects as inferred by the Vision-FM (Llama 3.2 90B Vision Instruct~\cite{grattafiori2024llama}) and verified by the operator. To evaluate the impact of semantic enhancement for planning, we compare the performance of 5 large language models: Llama 3.1 8B and 405B Instruct~\cite{grattafiori2024llama}, GPT 4o mini and o3 mini~\cite{gpt4}, and Gemma 3 27B and Gemini Flash 2.0~\cite{pichai2024gemini}, across three object retrieval scenarios: \textit{Get-Sponge}, \textit{Get-Coffee}, and \textit{Get-Tissue}. The \textit{Get-Sponge} task is an unambiguous reasoning instance where the target object (a sponge) is placed in its logically and semantically appropriate location—i.e., in the kitchen by the sink.
This makes it relatively straightforward for LLMs to infer that the kitchen sink is the obvious place to search for the sponge. In the \textit{Get-Coffee} task, we introduce a misleading association: the coffee powder is placed on a tray in the office rather than in the lounge or kitchen. This reflects a common real-world scenario where an object is not in a semantically expected location, requiring the agent to rely on a prior mapping phase to identify its placement. Finally, the \textit{Get-Tissue} task presents a many-to-one mapping scenario, where multiple tables (at least one per location) could plausibly contain a tissue box, but only one actually does. This setting reflects a frequent occurrence in household environments, where several semantically valid locations might exist for a given object. In such cases, an accurate map with the object's current location is essential for efficient task completion.
Queries are given in natural language, and the model is asked to reply with the skill sequence most likely to solve the task.

\subsection{Baseline Performance}


    
    

\begin{figure}[ht!]
    \centering
    \includegraphics[width=0.9\columnwidth]{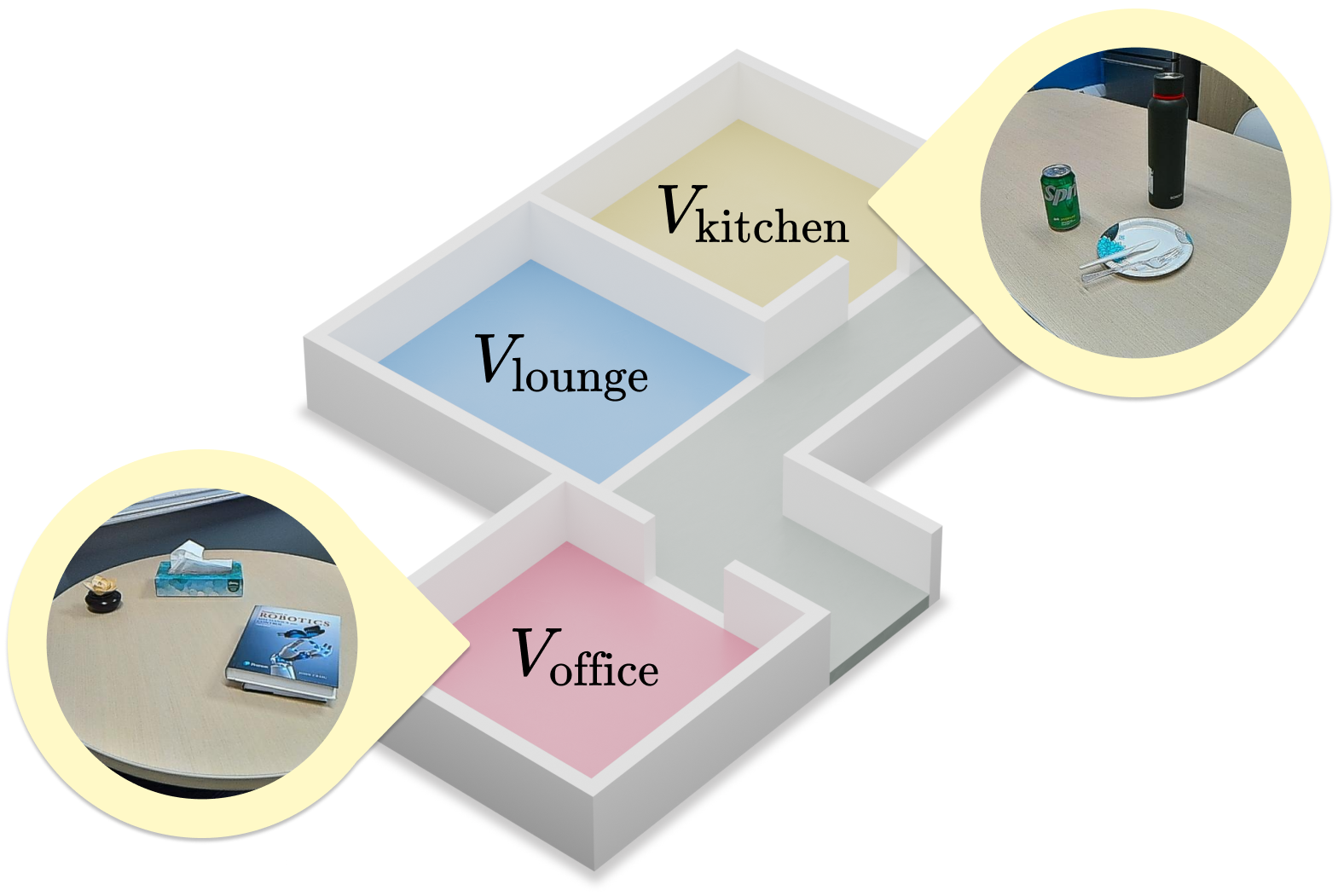}
    \caption{\small{\textbf{Semantic Ambiguity.} The topological map of our indoor environment contains several instances of drinks and tables, two nodes of which are pictured here. When given a task asking for a ``tissue'', each FM knows that a desk or table is a likely location for a tissue box, but is forced to make a guess without additional semantic context. Similarly, the FM must guess between drinks when queried for a beverage with no context of who a drink belongs to. }}
    \label{fig:qualitative}
\end{figure}

We define our baseline map as a Scene JSON with no semantic enhancement, meaning only top-level location information is provided (e.g., office desk, kitchen fridge), inspired by techniques such as~\cite{bumble}. This consists of our nine semantic nodes over three zones, but without object context.
As the baseline method lacks contextual information about the objects present, the LLM is forced to infer where an object may be located based on semantic cues commonly associated with each location. Usually in large indoor environments, multiple similar locations exist, introducing semantic ambiguity. For example, when asked to retrieve a tissue, the language model understands that tables are a common location for tissues to be located, but must guess between the various tables within the environment as pictured in Fig~\ref{fig:qualitative}, leading to inconsistent task performance. 
This trend is seen across all LLMs tested as shown in Table~\ref{tab:task_success}.

In case of indirect queries -- where the target object is not explicitly stated (e.g., \textit{``I have a cold and I'm feeling a bit sniffly.''}, tissue implied) -- the LLM was unable to answer correctly using the baseline map as shown in Table~\ref{tab:indirect}.


\subsection{Semantic Enhancement}
When provided with semantic context about the objects present at each node, all LLMs are able to correctly infer the target skill sequence even in adversarial cases like the \textit{Get-Coffee} or \textit{Get-Tissue} scenarios. The semantic context also allows even a relatively small LLM (like Gemma 3 27 B) to accurately plan the correct skill sequence even in cases of tricky indirect queries as shown in Table~\ref{tab:indirect}. We also explore queries that require knowledge of object ownership. By including two people, Bob and Alice, into the Scene JSON and tagging two of Bob's items, we can plan for tasks involving someone's items or current locations. While the FM can plan for direct object queries on the SENT-Map, lack of ownership and location tagging causes hallucinations in ambiguous scenarios, such as assuming anything near Bob's is owned by him. Results are shown in Table~\ref{tab:indirect}.




\section{Conclusion and Future Work}
We introduced SENT-Maps, Semantically-Enhanced Topological Maps for autonomous robot navigation and manipulation. SENT-Maps represent the environment in JSON format, enabling humans to edit and foundation models to parse the environment for downstream planning. We demonstrate a framework for constructing a SENT-Map using an operator guided mapping phase, and a planning phase, both assisted by foundation models. Through semantic enhancement, we showed foundation models were able to plan more effectively for locally-deployable (27B param) FMs and from indirect queries. 

An important limitation of this work is the effort required from an operator to attain sufficient semantic enhancement. This may be especially limiting in large or complex environments where the required operator effort may not scale well with the environment. Additionally, the SENT-Maps resulting from complex environments may result in long or complex JSON files that smaller LLMs could have trouble parsing correctly or that may confuse the human operator.  
To address these limitations, we hope to explore a tradeoff between operator-level semantic enhancement and scalable mapping methods for robots such as Hydra~\cite{hughes2022hydra}, as well as the semantic complexity tradeoff. We also hope to further investigate the ability for users to interpret and augment the SENT-Map, allowing them to alter the map through a user interface during and after mapping, as well as while planning.

{\small
	\bibliographystyle{IEEEtran}
	\bibliography{IEEEabrv,refs}

@article{bumble,
  title={BUMBLE: Unifying Reasoning and Acting with Vision-Language Models for Building-wide Mobile Manipulation},
  author={Shah, Rutav and Yu, Albert and Zhu, Yifeng and Zhu, Yuke and Mart{\'\i}n-Mart{\'\i}n, Roberto},
  journal={IEEE International Conference on Robotics and Automation (ICRA)},
  year={2025},
  organization={IEEE}
}

@article{KUIPERS199147,
title = {A robot exploration and mapping strategy based on a semantic hierarchy of spatial representations},
journal = {Robotics and Autonomous Systems},
volume = {8},
number = {1},
pages = {47-63},
year = {1991},
note = {Special Issue Toward Learning Robots},
issn = {0921-8890},
doi = {https://doi.org/10.1016/0921-8890(91)90014-C},
author = {Benjamin Kuipers and Yung-Tai Byun},
}

@article{hughes2022hydra,
    title={Hydra: A Real-time Spatial Perception System for {3D} Scene Graph Construction and Optimization},
    fullauthor={Nathan Hughes, Yun Chang, and Luca Carlone},
    author={N. Hughes and Y. Chang and L. Carlone},
    booktitle={Robotics: Science and Systems (RSS)},
    year={2022},
}

@ARTICLE{FM-Fusion,
  author={Liu, Chuhao and Wang, Ke and Shi, Jieqi and Qiao, Zhijian and Shen, Shaojie},
  journal={IEEE Robotics and Automation Letters}, 
  title={FM-Fusion: Instance-Aware Semantic Mapping Boosted by Vision-Language Foundation Models}, 
  year={2024},
  volume={9},
  number={3},
  pages={2232-2239},
  doi={10.1109/LRA.2024.3355751}
}

@INPROCEEDINGS{Slam++,
  author={Salas-Moreno, Renato F. and Newcombe, Richard A. and Strasdat, Hauke and Kelly, Paul H.J. and Davison, Andrew J.},
  booktitle={2013 IEEE Conference on Computer Vision and Pattern Recognition}, 
  title={SLAM++: Simultaneous Localisation and Mapping at the Level of Objects}, 
  year={2013},
  volume={},
  number={},
  pages={1352-1359},
  doi={10.1109/CVPR.2013.178}
}

@inproceedings{SemanticMapping,
author = {Zeng, Zhen and Zhou, Yunwen and Jenkins, Odest Chadwicke and Desingh, Karthik},
title = {Semantic Mapping with Simultaneous Object Detection and Localization},
year = {2018},
publisher = {IEEE Press},
doi = {10.1109/IROS.2018.8594205},
booktitle = {2018 IEEE/RSJ International Conference on Intelligent Robots and Systems (IROS)},
pages = {911–918},
numpages = {8},
location = {Madrid, Spain}
}

@misc{ClipField,
author = {Shafiullah, Nur and Paxton, Chris and Pinto, Lerrel and Chintala, Soumith and Szlam, Arthur},
year = {2022},
month = {10},
pages = {},
title = {CLIP-Fields: Weakly Supervised Semantic Fields for Robotic Memory},
doi = {10.48550/arXiv.2210.05663}
}

@INPROCEEDINGS{OpenVocab,
  author={Chen, Boyuan and Xia, Fei and Ichter, Brian and Rao, Kanishka and Gopalakrishnan, Keerthana and Ryoo, Michael S. and Stone, Austin and Kappler, Daniel},
  booktitle={2023 IEEE International Conference on Robotics and Automation (ICRA)}, 
  title={Open-vocabulary Queryable Scene Representations for Real World Planning}, 
  year={2023},
  volume={},
  number={},
  pages={11509-11522},
  doi={10.1109/ICRA48891.2023.10161534}
}

@misc{vlmaps,
      title={Visual Language Maps for Robot Navigation}, 
      author={Chenguang Huang and Oier Mees and Andy Zeng and Wolfram Burgard},
      year={2023},
      eprint={2210.05714},
      archivePrefix={arXiv},
      primaryClass={cs.RO}
}

@inproceedings{openfusion,
  author={Kashu Yamazaki and Taisei Hanyu and Khoa Vo and Thang Pham and Minh Tran and Gianfranco Doretto and Anh Nguyen and Ngan Le},
  title={Open-Fusion: Real-time Open-Vocabulary 3D Mapping and Queryable Scene Representation},
  year={2024},
  cdate={1704067200000},
  pages={9411-9417},
  booktitle={ICRA}
}

@misc{conceptgraph,
      title={ConceptGraphs: Open-Vocabulary 3D Scene Graphs for Perception and Planning}, 
      author={Qiao Gu and Alihusein Kuwajerwala and Sacha Morin and Krishna Murthy Jatavallabhula and Bipasha Sen and Aditya Agarwal and Corban Rivera and William Paul and Kirsty Ellis and Rama Chellappa and Chuang Gan and Celso Miguel de Melo and Joshua B. Tenenbaum and Antonio Torralba and Florian Shkurti and Liam Paull},
      year={2023},
      eprint={2309.16650},
      archivePrefix={arXiv},
      primaryClass={cs.RO}
}

@misc{clio,
      title={Clio: Real-time Task-Driven Open-Set 3D Scene Graphs}, 
      author={Dominic Maggio and Yun Chang and Nathan Hughes and Matthew Trang and Dan Griffith and Carlyn Dougherty and Eric Cristofalo and Lukas Schmid and Luca Carlone},
      year={2024},
      eprint={2404.13696},
      archivePrefix={arXiv},
      primaryClass={cs.RO}
}

@article{gpt4,
  title={Gpt-4 technical report},
  author={Achiam, Josh and Adler, Steven and Agarwal, Sandhini and Ahmad, Lama and Akkaya, Ilge and Aleman, Florencia Leoni and Almeida, Diogo and Altenschmidt, Janko and Altman, Sam and Anadkat, Shyamal and others},
  journal={arXiv preprint arXiv:2303.08774},
  year={2023}
}

@article{grattafiori2024llama,
  title={The llama 3 herd of models},
  author={Grattafiori, Aaron and Dubey, Abhimanyu and Jauhri, Abhinav and Pandey, Abhinav and Kadian, Abhishek and Al-Dahle, Ahmad and Letman, Aiesha and Mathur, Akhil and Schelten, Alan and Vaughan, Alex and others},
  journal={arXiv preprint arXiv:2407.21783},
  year={2024}
}

@misc{pichai2024gemini,
  title={Introducing Gemini 2.0: our new AI model for the agentic era},
  author={Pichai, Sundar and Hassabis, D and Kavukcuoglu, K},
  year={2024},
  publisher={Google}
}

@inproceedings{3dllm,
 author = {Hong, Yining and Zhen, Haoyu and Chen, Peihao and Zheng, Shuhong and Du, Yilun and Chen, Zhenfang and Gan, Chuang},
 booktitle = {Advances in Neural Information Processing Systems},
 editor = {A. Oh and T. Neumann and A. Globerson and K. Saenko and M. Hardt and S. Levine},
 pages = {20482--20494},
 publisher = {Curran Associates, Inc.},
 title = {3D-LLM: Injecting the 3D World into Large Language Models},
 volume = {36},
 year = {2023}
}
}


\end{document}